\documentclass[letterpaper, 10 pt, conference]{ieeeconf}  % Comment this line out if you need a4paper

\IEEEoverridecommandlockouts                              % This command is only needed if 
\usepackage{mathptmx} % assumes new font selection scheme installed
\usepackage{times} % assumes new font selection scheme installed
\usepackage{amsmath,bm} % assumes amsmath package installed
\usepackage{amssymb}  % assumes amsmath package installed
\usepackage{graphicx}
\usepackage{comment}
\usepackage{caption}
\usepackage{mathtools}
\usepackage{upgreek}
\usepackage{gensymb}
\usepackage{comment}
\usepackage{algorithm}
\usepackage{algorithmic}
\usepackage[noadjust]{cite}
\usepackage{placeins}
\usepackage{xcolor}
\usepackage{url}
\usepackage{mwe}
\usepackage{balance}
\usepackage{multirow}
\usepackage{array}
\usepackage{booktabs}
\usepackage{pifont}
\usepackage{float}
\usepackage{hyperref}

\DeclarePairedDelimiter\abs{\lvert}{\rvert}%
\DeclarePairedDelimiter\norm{\lVert}{\rVert}%

\makeatletter
\let\oldabs\abs
\def\abs{\@ifstar{\oldabs}{\oldabs*}}
\let\oldnorm\norm
\def\norm{\@ifstar{\oldnorm}{\oldnorm*}}
\makeatother

\newcommand{\R}{\mathbb{R}}
\newcommand{\bR}{\mathbf{R}}
\newcommand{\bp}{\mathbf{p}}
\newcommand{\bT}{\mathbf{T}}
\newcommand{\ve}[1]{\mathbf{#1}}

\title{\LARGE \bf
The N2D Haptic Glove: A Multi-Finger Glove for 2D Directional Force Feedback for Contact Rich Manipulation}

\author{
Yao-Ting Huang$^{\dagger,1}$, Jake Honma$^{\dagger,2}$,
Omar Hernandez$^{\dagger,1}$, Logan Li$^{\dagger,3}$, Kaitlin Calimbahin$^{\dagger,1}$, Bryce Hackel$^{1}$, 
\\
Michael C.\ Yip$^{1}$, \textit{Senior Member, IEEE}
\thanks{$^\dagger$ Equal contribution.}
\thanks{$^1$ Electrical and Computer Engineering, University of California San Diego, La Jolla, CA 92093, USA.}
\thanks{$^2$ Mechanical and Aerospace Engineering, University of California San Diego, La Jolla, CA 92093, USA.}\thanks{$^3$ Bioengineering, University of California San Diego, La Jolla, CA 92093, USA.}
}
\begin{document}

\maketitle
\thispagestyle{empty}                                          
\pagestyle{empty}

%%%%%%%%%%%%%%%%%%%%%%%%%%%%%%%%%%%%%%%%%%%%%%%%%%%%%%%%%%%%%%%%%%%%%%%%%%%%%%%%

\begin{abstract}
% Humans rely on directional fingertip cues, both normal and tangential forces, to probe, sense slip, and control shear during manipulation, yet most wearable gloves render only vibration or normal force, which leaves direction ambiguous. Without directional cues, users must infer contact direction from vision, which leads to over-gripping, dropped objects, and slower, less precise control in virtual reality (VR) simulations and robotic teleoperation. We present the N2D Haptic Glove, a multi-finger wearable device that renders planar vector forces at each fingertip using a compact, transparency-focused mechanical design with on-hand actuation that preserves natural motion. 
% %In evaluation, the glove maintained nearly the full range of motion across various user hand sizes and produced fingertip forces that closely matched commanded vectors. Experiments involving VR slip object handling, a peg-in-hole task, and teleoperated push-button probing showed that directional feedback reduced completion time and errors while improving workload and confidence compared to visual-only and one-dimensional haptic feedback. 
% Our findings demonstrate that multi-directional fingertip forces are critical for effective haptic interaction, establishing the N2D Haptic Glove as a transformative device for the future of VR simulations, teleoperation, and robot learning.

Humans rely on directional fingertip forces to probe and regulate contact during manipulation, yet most wearable haptic gloves render only vibration or single-axis force, leaving force direction ambiguous. Without directional cues, users must infer contact force from vision alone, often leading to over-pressing, inconsistent control, and reduced precision in robotic teleoperation. We present the N2D Haptic Glove, a multi-finger wearable device that renders planar flexion-extension fingertip forces using capstan-drive transmissions for high-transparency force feedback. Through benchtop validations and a user study involving haptic teleoperation of a robotic arm and hand, we demonstrate that compared to visual-only and single-axis haptic baselines, planar fingertip feedback significantly reduces contact force error during precise manipulation, improves trial-to-trial consistency, and enhances overall user experience in axial probing tasks. These findings establish the N2D Haptic Glove and directional finger-based haptics devices as a promising modality for contact-rich teleoperation, immersive virtual reality simulations, and robot learning from demonstrations. N2D Haptic Glove’s hardware and software system will be fully
open-sourced at \href{https://ucsdarclab.github.io/n2d-glove/}{https://ucsdarclab.github.io/n2d-glove/}.
\end{abstract}

\section{Introduction}
%Haptic devices have been utilized in various immersive contexts, ranging from interacting with virtual reality (VR) environments in simulation to teleoperating robots for minimally invasive surgery \cite{vdMeijden09}. Moreover, recent interest in using human demonstrations to train robots involves human teleoperators controlling manipulators within real or virtual worlds, where haptic feedback has been shown to collect tighter and more informed distributions of trajectories for robot learning \cite{cuan2024leveraging}. %

Haptic devices have been widely utilized in immersive contexts, ranging from interacting with virtual reality (VR) environments to teleoperating robotic surgical systems for minimally invasive surgery. Recent interest in using human demonstrations to train robot imitation learning policies requires human teleoperators to control manipulators within real or virtual worlds, where haptic feedback has been shown to collect tighter and more informed distributions of trajectories \cite{cuan2024leveraging}.

%Most current haptic devices primarily provide proprioceptive feedback to the hand, with most interfaces involving the user hand grasping a handle \cite{choi2018claw} or stylus \cite{touch3stylus}. Other forms of haptic feedback exist that involve vibrotactile sensations, but do not provide the necessary kinesthetic feedback for force rendering \cite{arasan2013haptic}. %

%One exciting area with limited exploration is multi-fingered haptic feedback. Providing force feedback independently to each finger can be especially beneficial for the operation of humanoid robots with multiple fingers, which is rapidly increasing in popularity for research and industry \cite{tong2024advancements}. However, while a variety of methods currently exist to provide haptic feedback to fingers, from tendon-driven gloves \cite{baik2020haptic} to finger-mounted vibration pads \cite{trinitatova2025fidtouch}, an important limitation remains - the lack of comprehensive fingertip haptics capable of reproducing both normal and tangential forces.%

% With the rise of the humanoid form factor for robotics, an emerging challenge is effective control methods for robotic hands. Recent techniques like VR controller teleoperation suffer from a lack of feedback cues and can result in slow and imprecise control. A workaround is implementing alternative cues to overcome this information gap, such as vibration motors, pneumatic grids, and cable-driven mechanisms. %

With the rise of the humanoid form factor for robotics, a key challenge has been developing effective control methods for robotic hands. While recent VR-based teleoperation systems enable dexterous control through visual feedback \cite{cheng2024opentv}, the absence of haptic cues limits operators' ability to regulate contact forces and avoid errors \cite{nitsch2013meta}. To address this, existing haptic gloves provide cable-driven flexion resistance \cite{baik2020haptic, zhang2025doglove, senseglover1} and fingertip-mounted cutaneous feedback \cite{trinitatova2025fidtouch, haptx2025}.

% ----- Start Figure 1 -----
\begin{figure}[t!]
    \vspace{-0.5em} % reduce space above
    \centering
    \includegraphics[width=.95\linewidth]{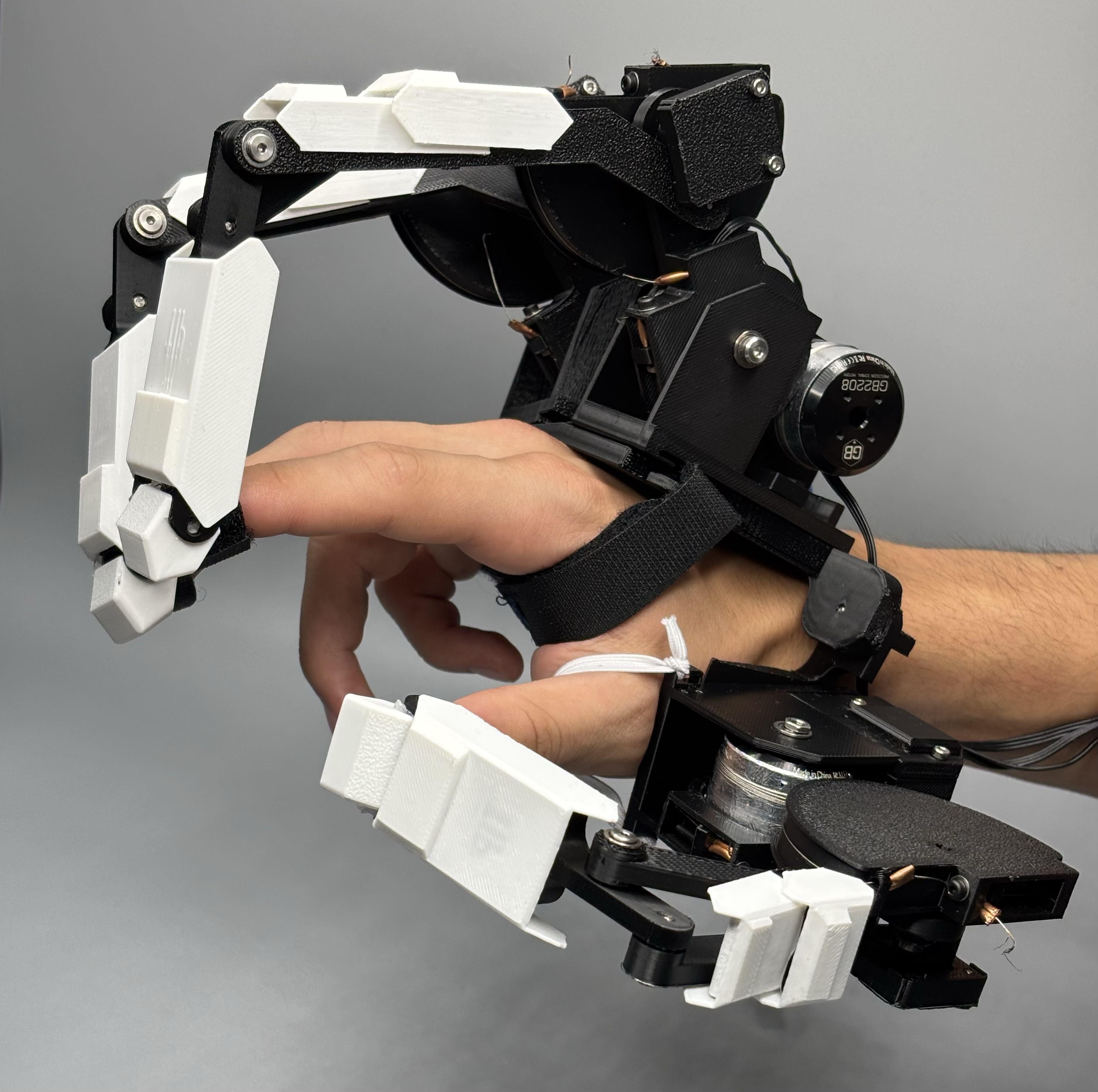}
    %\vspace{-1.5em}   % reduce space between image and caption
    \caption{\textbf{The N2D Haptic Glove:} is the first, multi-fingered glove design that offers 2D force feedback along the axial direction of multiple fingers. The glove provides multi-directional haptic rendering, an integral feedback capability for tasks requiring probing, gentle grasping, slip detection, etc., which rely on directional forces.}
    \label{fig:intro_figurel}
    \vspace{-1.5em}   % reduce space after figure
\end{figure}
% ----- End Figure 1 -----

Directional haptic rendering in finger interactions remains underdeveloped. In this work, we define \emph{transverse} forces as those aligned with finger flexion/extension at the finger pad, and \emph{axial} forces as those acting along the length of the finger, such as during poking. Consider moving a finger through a cluttered set of objects: the perception of axial forces at the finger tip can be just as vital — if not more so — than transverse forces generated along the closing grasping direction. Accurate perception of both is essential for contact-rich manipulation tasks, including probing, gentle grasping, and slip detection.

%Old ending to paragraph above: Measurements such as slip and axial probing force cannot be provided to the fingers without rendering directional force.

A current indirect approach to approximate directional feedback is to use skin-shear on a held device \cite{schorr2015tactor}, \cite{gwilliam2013haptic}; however, delivering axial force feedback directly to individual fingertips is arguably more natural and interpretable.

In this paper, we present the first wearable haptic interface with on-hand actuation capable of rendering planar kinesthetic forces at multiple finger tips (Fig.~\ref{fig:intro_figurel}). Named the N2D Haptic Glove — for $N$ fingers and 2D force rendering per finger — doubling the DOF per finger of active force-feedback over existing designs, with applications in robot teleoperation, VR simulations, and imitation learning.

% ----- Starts of Table I -----
{\small
\begin{table*}[h]
    \vspace{2.0mm} % reduce space above the table
    \centering
    \caption{Comparison of Fingertip-Feedback Systems}
    \label{tab:glove_comparison}
    \vspace{-0.5em} % reduce space between caption and table
    \renewcommand{\arraystretch}{1}
    \setlength{\tabcolsep}{5pt}

    \newcommand{\cmark}{\checkmark}
    \newcommand{\xmark}{\ding{55}}

    \small
    \resizebox{\textwidth}{!}{%
    \begin{tabular}{@{}clccccccc@{}}
        \toprule
        \# & Feature & SenseGlove R1 & HaptGlove & DOGlove & NURing & Fluid Reality & AirPush & N2D Glove \\
        \midrule
        1 & Multi-DOF Fingertip Forces  & \xmark & \xmark & \xmark & \cmark & \xmark & \cmark & \textbf{\cmark} \\
        2 & Kinesthetic Force Feedback    & \cmark & \cmark & \cmark & \cmark & \xmark & \xmark & \textbf{\cmark} \\
        3 & Multi-Digit (simultaneous)    & \cmark & \cmark & \cmark & \xmark & \cmark & \xmark & \textbf{\cmark} \\
        \bottomrule
    \end{tabular}%
    }

    \vspace{-1.5em} % reduce space after the table
\end{table*}
}

The key novel contributions of this paper are as follows:
\begin{enumerate}
    \item The first multi-fingered haptic glove capable of delivering planar force feedback to the finger,
    \item Design of the N2D Haptic Glove to achieve high transparency through low-friction, zero-backlash transmissions for clean, accurate haptic rendering,
    \item Validation of 2D planar directional feedback ability with benchtop testing, and 
    \item Demonstrating the benefits of N2D Haptic Glove’s directional haptics in a teleoperation user study.
\end{enumerate}
% ----- End of Table I -----
\section{Related Works}

In the realm of fingertip haptics, existing wearable interfaces can be grouped by the combination of features they support: multi-DOF fingertip forces, kinesthetic force feedback, and multi-digit actuation.

%\subsection{Paragraph 1 - Directional but single-finger}
Several devices render controllable force directions at the fingertip but are generally limited to single-finger designs. Skin-stretch devices generate tangential cues through multi-directional deformation of the fingerpad \cite{trinitatova2025fidtouch}, \cite{quek2015sensory}, \cite{min2025ultralight}. AirPush demonstrates multi-directional fingertip forces using compressed air with 2-DOF control \cite{ma2024airpush}. Similarly, the NURing delivers two-dimensional kinesthetic deflection cues via tendon-driven actuation on the index finger \cite{trzpit2025nuring}. These designs achieve directional rendering but cannot scale to multi-finger use cases such as dexterous grasping.

%\subsection{Paragraph 2 - Multi-finger but cutaneous-only}
A second class of devices provides feedback across multiple fingers but is limited to cutaneous stimulation without kinesthetic force. Vibrotactile actuators simulate contact events and palpation cues through fingertip-mounted vibration pads \cite{kim2018hapcube}, \cite{pacchierotti2015cutaneous}. Fluid Reality achieves high-resolution pressure rendering across five fingertips using pump arrays \cite{shen2023fluid}. These approaches offer rich tactile information but cannot reproduce the sustained, directional forces needed for precise manipulation.

%\subsection{Paragraph 3 - Multi-finger but single-axis}
Several integrated glove systems target multi-finger kinesthetic feedback. SenseGlove R1 delivers four-finger resistance through an active force-feedback exoskeleton \cite{senseglover1}. Dexmo provides multi-digit resistive feedback through a link-bar exoskeleton \cite{gu2016dexmo}. DOGlove provides five-finger active force feedback augmented with vibrotactile sensing at a low cost \cite{zhang2025doglove}. HaptGlove combines pneumatic clutches with skin indenters across all five fingers \cite{qi2023haptglove}. While each system delivers active kinesthetic force feedback across multiple fingers, all restrict fingertip forces to a single axis without directional rendering.

% Closing:
Collectively, these systems represent significant progress across individual dimensions of haptic feedback, but none combine directional fingertip forces, kinesthetic resistance, and multi-digit coverage in a single device. The N2D Haptic Glove addresses this gap (Table~\ref{tab:glove_comparison}).

% ----- End of Topic Sentence 3 -----

\section{Methods}

The following sections describe: (A) the mechanical design of the N2D Haptic Glove linkage system, (B) the electromechanical framework, (C) kinematic analysis of the glove, and (D) a 2D planar haptic feedback rendering approach for multi-fingered force reflectance.

% --- Previous Method 1A Section Bullet Points ---
% \textbf{The design of the haptic glove is based on the biomechanical degrees of freedom of the human hand.} 
% \begin{itemize}
% \item (explain breaking hand down into planar motions, adduction and abduction), etc. 
% \item 
% \end{itemize}

% --- Start of Methods Topic Sentence 1A ---
\subsection{Mechanical Design Overview}
% The design of the haptic glove is based on allowing unrestricted natural motion of the finger joints, governed by the biomechanical DOFs of the human hand. Following established biomechanical models, hand movement is often simplified to planar motions of flexion and extension, which occur at the metacarpophalangeal (MCP), interphalangeal (PIP), and distal interphalangeal (DIP) joints of the fingers, together with adduction and abduction, which describe the lateral spreading and closing of the fingers relative to the middle of the hand. While most fingers primarily operate through flexion and extension planar motions, the thumb is unique in also enabling adduction and abduction at the trapeziometacarpal joint (TMCJ) -- a movement essential for precision when grasping objects. Alongside the thumb and index finger, the middle finger provides additional stability and support during grasping, making these three digits central to most everyday picking up and probing tasks. 

The N2D Haptic Glove is designed to preserve the natural biomechanical DOFs of the hand. Following established models \cite{fok2010development}, the MCP joint is treated as a saddle joint allowing flexion-extension and adduction-abduction, while the PIP and DIP joints are modeled as revolute joints constrained to flexion-extension. The thumb additionally performs opposition at the TMC joint. The thumb, index, and middle fingers are central to most grasping and probing tasks due to their roles in precision and stability.

% --- End of Methods Topic Sentence 1A ---
% --- Previous Method 2A Section Bullet Points ---
% \textbf{The haptic glove comprises multiple directional haptic subsystems per finger, articulating on a skeleton backbone.}
% \begin{itemize}
% \item Each finger receiving force feedback engages with a linkage system providing 2DOF, and with $n$ fingers supported, in which the N2D gets its name. 
% \item Each linkage system is secured on a passive rotary joint that provides adduction/abduction between fingers. The thumb requires additional degrees of freedom for free movement, and thus includes an additional passive rotary joint. 
% \item Figure \ref{fig:mechatronics_overview}
% \item The directional force linkage design on each finger is achieved using a parallel mechanism, with planar control achieved by rotating the proximal linkages using a capstan cable-drive (further details in the kinematic modeling section).
% \end{itemize}

\begin{figure}
    \vspace{2mm} % reduce space above
    \centering
    \includegraphics[width=0.95\linewidth]{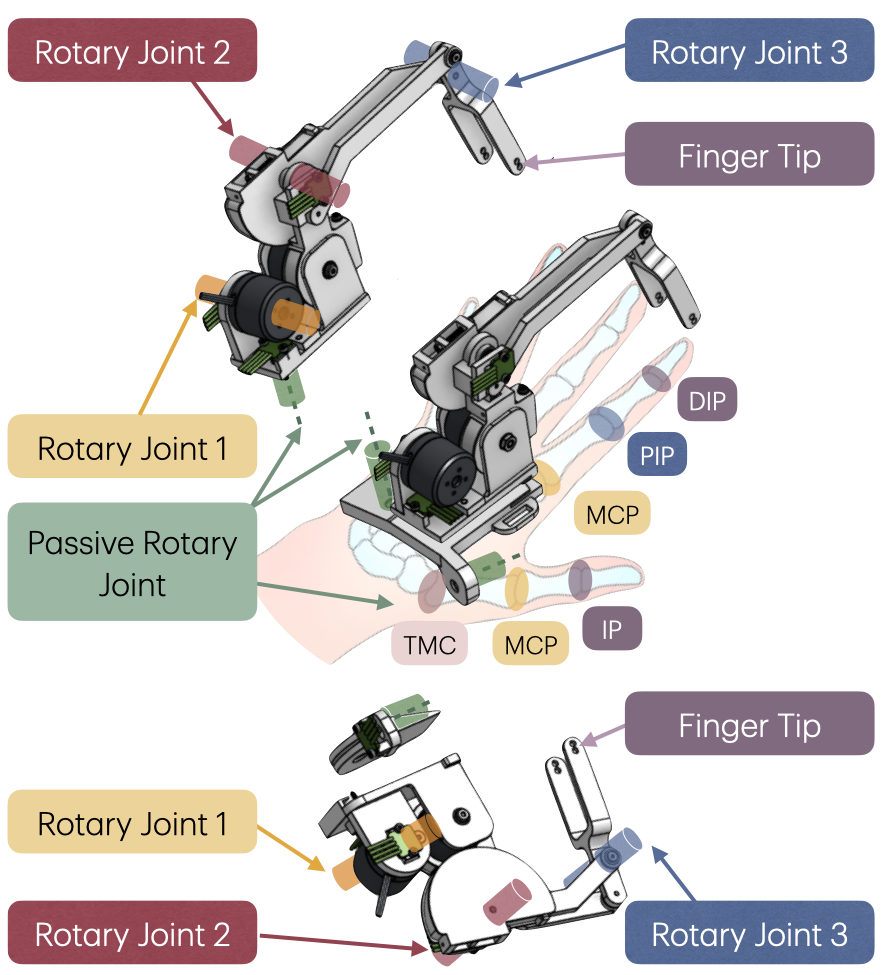}
    %\vspace{-0.5em} % reduce space between image and caption
    \caption{Deconstructed view of the N2D Haptic Glove showing finger linkage joint layout and its correspondence to hand anatomy.}
    \label{fig:mechatronics_overview}
    \vspace{-1.5em} % reduce space after figure
\end{figure}

% --- Start of Methods Topic Sentence 2A ---
The device consists of three modular finger subsystems mounted along a structural backbone. Each subsystem uses a crossed four-bar linkage with an additional link to permit full flexion-extension while enabling 2D planar force feedback. Planar fingertip forces are generated by actuating the proximal links via capstan cable drives (see Kinematic Modeling). Each linkage is mounted on a passive rotary joint for finger abduction-adduction, with the thumb having an additional passive joint for TMC opposition. The resulting 2-DOF planar formulation provides physiologically grounded kinematics while enabling controlled directional force feedback at each fingertip (Fig.~\ref{fig:mechatronics_overview}), consistent with validated models of lateral pinch \cite{lemos2024biomechanical}.

%This architecture reflects established biomechanical models: the MCP joint is treated as a saddle joint allowing flexion–extension and adduction–abduction, while the PIP and DIP joints are modeled as rotary joints constrained to flexion–extension \cite{fok2010development}. The resulting 2-DOF planar formulation provides physiologically grounded kinematics while enabling controlled directional force feedback at each fingertip (Fig.~\ref{fig:mechatronics_overview}), consistent with validated computational models of lateral pinch \cite{lemos2024biomechanical}.%

%This structure aligns with biomechanical models of the finger, where the MCP joint is treated as a saddle joint allowing flexion–extension and adduction–abduction, and the PIP and DIP joints are modeled as rotary joints constrained to flexion–extension \cite{fok2010development} but also enables controlled directional force feedback on each finger, as illustrated in Figure~\ref{fig:mechatronics_overview}. This reduction to a 2-DOF planar model for force feedback offers a practical yet physiologically accurate framework for hand kinematics, and has been validated in computational simulations of lateral pinch that replicate physiological joint behavior with high fidelity \cite{lemos2024biomechanical}.

 % For the figure, let's have the N2D Haptic Glove over the labeled anatomy of the hand

% --- End of Methods Topic Sentence 2A ---
Beyond directional force feedback, the glove prioritizes haptic transparency. All active DOFs are actuated via capstan drives powered by brushless motors, avoiding geared transmissions that introduce backlash, friction, and damping. Actuators are mechanically grounded at the base of the linkage system, preventing motor masses from rotating about the finger joints and eliminating the sensation of actuator mass swinging on the finger. All transmissions remain on the hand, avoiding the resistance, hysteresis, and unmodeled dynamics introduced by off-hand routing such as Bowden cables or pneumatic hoses.

%In addition to enabling directional force feedback per finger, a key design feature is that haptic transparency is prioritized in the haptic glove design. Friction and damping caused by non-ideal transmissions (using gears, servomotor gearing stages, belt drives) can result in non-negligible clouding of haptic signal-to-noise-ratio (SNR), and thus negatively affect the wearer's experience. Thus, direct-drive ungeared motors with capstan drives are used for all active DOF. Motors are furthermore fixed to the base of the linkage system (no flying motors) such that their centers of mass do not revolve around axes of rotation during finger flexion-extension. Therefore, the strenuous feeling of the motors inertia swinging on the user's hand is avoided. In short, all mechanical considerations (cable-driven, ungeared motors, no flying motors, linkage systems over tendons and tendon routings, bearing-supported axles) were chosen for high haptic transparency for the N2D glove.  

%A key design choice is on-hand actuation — all mechanical transmissions remain on the hand. This avoids the resistance, hysteresis, and unmodeled dynamics introduced by off-hand routing such as Bowden cables or pneumatic hoses.%

The N2D glove weighs 562 g, with a detailed breakdown provided in Table \ref{tab:components_weight}; the battery is excluded, as it is off-hand. Nearly half of the total mass comes from PLA components, suggesting that reduced infill, structural redesign, and lighter materials could substantially decrease weight. The other major contribution comes from the gimbal motors, which is a notable design constraint. Future advances in gimbal motor miniaturization will also lower overall system mass.

% --- Previous Method 1B Section Bullet Points ---
% \subsection{Electromechanical Framework}
% \textbf{The electromechanical architecture combines direct-drive GB2208 outrunner gimbal motors with DRV8313 brushless DC motor drivers and AS5048B magnetic encoders for frictionless torque transparency.}
% \begin{itemize}
% \item Each combination of motor-sensor pair talks to Teensy 4.1 microprocessor board, providing 10 Hz control loop to support responsive haptic feedback. %TODO: Figure out frequency for control loop
%     % Frequency depends on the frequency between topics in ROS
%     \item ...
% \end{itemize}

% The overall control block diagram for the entire electromechanical architecture is provided in Figure \ref{fig:control_block_diagram}.
% \begin{itemize}
% \item
% \end{itemize}
% \begin{figure*}
%     \centering
%     \includegraphics[width=\linewidth]{Figures/system_diagram.pdf}
%     \caption{\textbf{Control block diagram showing the system architecture.} The $N$ joints run independent torque control, serially communicating with a common microcontroller that performs kinematic and haptic rendering mappings from Cartesian space to joint space and serially communicating with a PC. Initialization of kinematics is required for accurate directional force rendering to account for hand sizes.}
%     \label{fig:control_block_diagram}
% \end{figure*}

\begin{figure}[t]
    \centering
    \vspace{2mm}
    \includegraphics[width=0.8\linewidth]{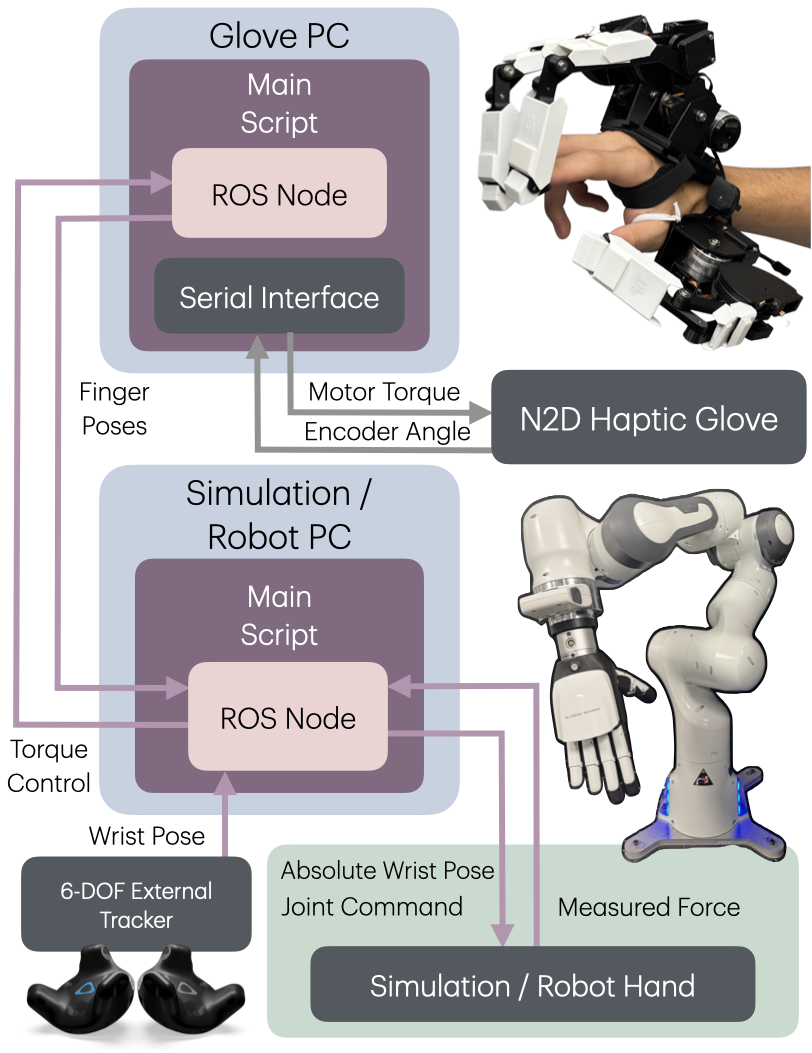}
    \caption{System architecture of the N2D Haptic Glove. The Glove PC computes finger poses from encoder data and sends joint commands to the Simulation/Robot PC via ROS. Sensed interaction forces are returned and mapped to motor commands.}
    \label{fig:control_block_diagram}
    \vspace{-1.5em}
\end{figure}

% --- Start of Methods Topic Sentence 2A ---
\subsection{Electromechanical Architecture}
%Figure~\ref{fig:control_block_diagram} summarizes the electromechanical stack and distributed control that enable torque-transparent direct-drive actuation in the N2D Haptic Glove. Direct-drive GB2208 gimbal motors were chosen as they are currently the smallest, commercially available gimbal motor (high-torque output, low speed) without planetary gearing. The six motors are powered by DRV8313 three-phase drivers from an 14\,V supply and each instrumented with a AS5048B 14-bit absolute magnetic encoders for rotor angle feedback. Four additional AS5048B magnetic encoders are used to monitor the four passive abduction-adduction joints. Each motor–encoder pair interfaces with a Teensy 4.1 microcontroller executing field-oriented control (FOC) in torque mode at \textit{{1}\,kHz}, while higher-level torque commands are sent at \textit{{10}\,Hz} from the Glove PC over a serial connection.

Fig.~\ref{fig:control_block_diagram} summarizes the electromechanical architecture and distributed control enabling torque-transparent capstan-drive actuation in the N2D Haptic Glove. GB2208 gimbal motors were selected for their compact form factor and high-torque, low-speed output without planetary gearing. Six motors are powered by DRV8313 three-phase drivers from a 12\,V supply, each paired with an AS5048B 14-bit absolute magnetic encoder for rotor angle feedback. Four additional AS5048B encoders measure the passive abduction–adduction joints.

Each motor–encoder pair interfaces with a Teensy 4.1 microcontroller running field-oriented control in torque mode at \textit{{1}\,kHz}, while higher-level torque commands are transmitted from the Glove PC at \textit{{10}\,Hz} over a serial connection.

% ---- Start of Table II ----
{\small
\begin{table}[t]
  \captionsetup{skip=6pt}  % space between caption and table body
  \vspace{1em}          % space before the table
  \footnotesize

  \centering
  \caption{Weight breakdown of the N2D Haptic Glove}
  \label{tab:components_weight}
  \vspace{-0.5em} % reduce space between caption and table

  % shrink font and spacing a little
  \renewcommand{\arraystretch}{1}
  \setlength{\tabcolsep}{4pt}

  % resize to fit column width
  \resizebox{\columnwidth}{!}{%
    \begin{tabular}{@{}cp{3.0cm}cc@{}}
      \toprule
      \textbf{Qty} & \textbf{Component} & \textbf{Unit Weight (g)} & \textbf{Total (g)} \\
      \midrule
      6   & Gimbal Motor        & 39.5 & 237 \\
      6   & Motor Driver        & 4.5 & 27 \\
      10  & Encoder             & 1 & 10 \\
      1   & Main PCB            & 14 & 14 \\
      —   & PLA \& COTS parts & 274 & 274 \\
      \midrule
      \multicolumn{3}{r}{\textbf{Total Weight:}} & \textbf{562} \\
      \bottomrule
    \end{tabular}
  }
  \vspace{-2.1em} % space after the table
\end{table}}

On the Glove PC, a calibration script generates a user-specific profile of finger link lengths and 6-DOF finger poses. The main control script loads this profile and publishes hand state through a ROS node. An external VIVE Tracker 3.0 provides absolute wrist pose, which is fused with fingertip pose estimates. A corresponding ROS node on the robot side receives these poses and commands the robot arm and hand. Measured interaction forces are returned through ROS and mapped to torque commands for the glove, closing the bilateral control loop.

%To assign appropriate haptic feedback for each finger, either through contact detection in VR simulation or force measurements from a robotic hand, each finger’s kinematic model first computes fingertip position and forces relative to its base joint. These fingertip poses are transformed into a unified hand coordinate frame to determine their positions relative to the hand, after which the hand pose is estimated in the world frame to localize each digit within the environment. To describe the kinematics of the overall system, we will define the device as a collection of modular finger linkage subsystems fixed on a common base (Fig. \ref{fig:kinematics}). Each link of the device is defined as a rigid body with orientation $\bR \in SO(3)$ and position $\bp\in\R^3$. A transformation between bodies is represented by a homogeneous transform

\subsection{Kinematic Modeling}
Each finger's kinematic model computes fingertip pose and force relative to its base joint. These are transformed into a unified hand frame, then into the world frame using the estimated hand pose, localizing each digit in the environment.

\begin{figure}
    \centering
    \vspace{0.5em}
    \includegraphics[width=0.9\linewidth,keepaspectratio]{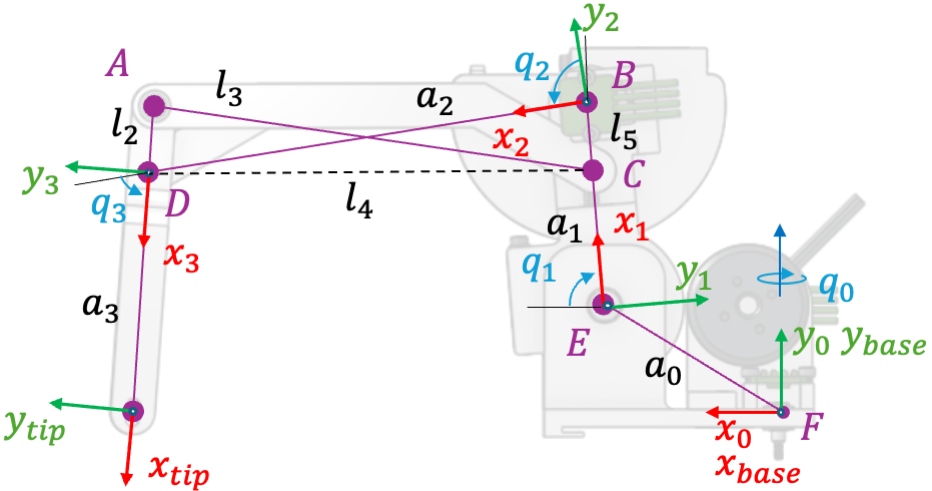}
    \includegraphics[width=0.9\linewidth,keepaspectratio]{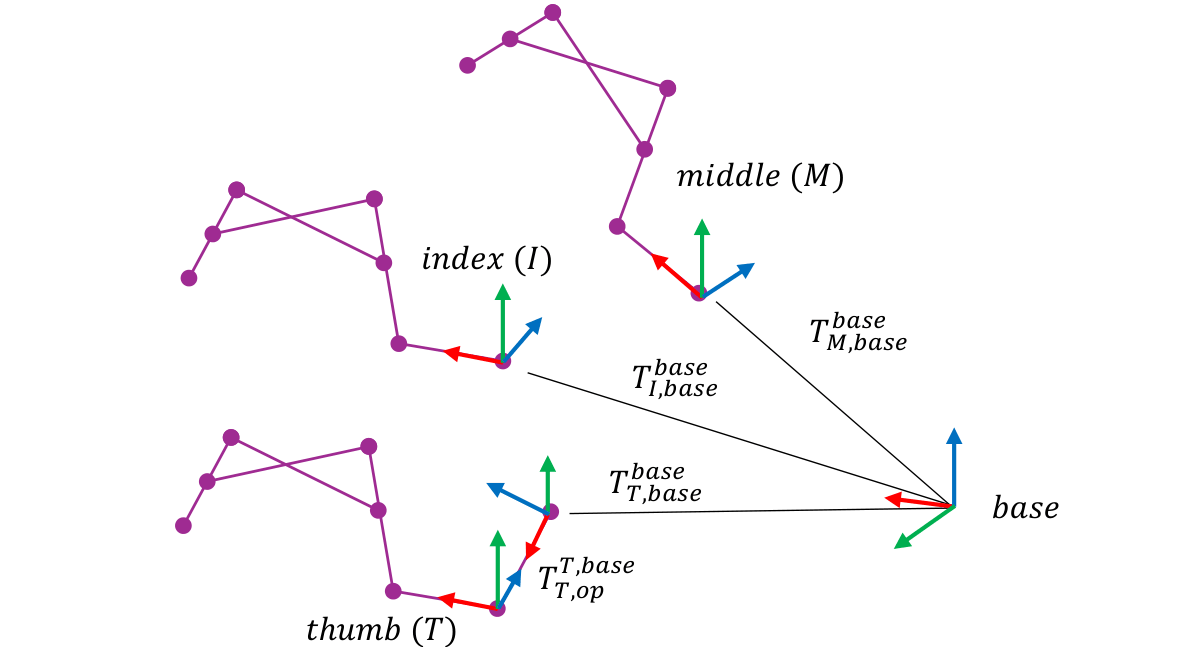}
    \caption{Kinematic analysis of a finger linkage. (Top) Single finger linkage with three DOFs defined by $q0, q1, q2$ (with $q3$ being a driven joint by $q1$ and $q2$) is shown. (Bottom) Finger linkages on the glove body show the overall coordinate frames with extra passive DOF for thumb opposition.}
    \label{fig:kinematics}
    \vspace{-1.5em}
\end{figure}

The device is modeled as a set of modular finger linkage subsystems mounted to a common base (Fig. \ref{fig:kinematics}). Each link is treated as a rigid body with orientation $\bR \in SO(3)$ and position $\bp \in \mathbb{R}^3$, and transformations between bodies are represented using homogeneous transforms:

\begin{equation}
\bT_i^j = \begin{bmatrix}\bR_i^j & \bp_i^j \\ \mathbf{0}_{1\times 3} & 1\end{bmatrix}\!
\end{equation}
which represents the pose of coordinate frame $i$ expressed in coordinate frame $j$.

We define a canonical base frame for the device $\{base\}$ with fixed origin set at the position-tracker's origin and orientation. Each digit, $f\in\{\text{T,I,M}\}$ (Thumb, Index, Middle) has a constant finger base coordinate frame $\bT_{T,base}^{base}, \bT_{I,base}^{base}, \bT_{M,base}^{base}$ described in the position-tracker origin. 

For non-thumb fingers ($f={I,M}$), the kinematics of the linkages from the finger's base ${f,base}$ to the tip can be described using three revolute joints $q_{f,0}, q_{f,1}, q_{f,2}$ and link lengths $a_{f,0}, a_{f,1}, a_{f,2}, a_{f,3}$. The distal tip pose $\{tip\}$ described in the base frame $\{base\}$ is 
{\small
\begin{multline}
\bT^{f,base}_{f,tip}(\ve{q}^{f},\ve{a}^{f}, \ve{l}^f, \ve{p}^f) =
\bT^{f,base}_{f,0}\;
\bT^{f,0}_{f,1}\;
\bT^{f,1}_{f,2}\;
\bT^{f,2}_{f,3}\;
\bT^{f,3}_{f,tip} = \\
\begin{bmatrix}
    \bR_y(q_{f,0}) & \mathbf{0}_{3\times 1} \\ \mathbf{0}_{1\times 3} & 1
\end{bmatrix}
\begin{bmatrix}
    \bR_z(q_{f,1}) & \bp_{f,1}^0 \\ \mathbf{0}_{1\times 3} & 1
\end{bmatrix}\\
\begin{bmatrix}
    \bR_z(q_{f,2}) & \mathrm{tr}_x(a_{f,1}+l_5) \\ \mathbf{0}_{1\times 3} & 1
\end{bmatrix}
\begin{bmatrix}
    \bR_z(q_{f,3}) & \mathrm{tr}_x(a_{f,2}) \\ \mathbf{0}_{1\times 3} & 1
\end{bmatrix}\\
\begin{bmatrix}
    \mathbf{I}_{3\times 3} & \mathrm{tr}_x(a_{f,3}) \\ \mathbf{0}_{1\times 3} & 1
\end{bmatrix}
\label{eq:FK-finger}
\end{multline}}

\noindent where $\bR_z$ is the $3\times 3$ rotation matrix about $z$ and $\mathbf{tr}_x(a_{(\cdot)})=[a_{(\cdot)}, 0, 0]^\top$. Angle $q_{f,3}$ is driven by $q_{f,1}$ and $q_{f,2}$. We solve $q_{f,3}$ via the cosine law by first computing the diagonal $CD$, then the internal angles $\angle ADC$ and $\angle BDC$, and finally taking their difference. Such that
{\small
\begin{align}
    CD &= l_4 = \sqrt{\,a_2^2 + l_5^2 + 2 a_2 l_5 \cos q_{f,2}} \\[6pt]
    \angle ADC &= \arccos\!\left( \frac{l_2^2 + l_4^2 - l_3^2}{2 l_2 \, l_4} \right)\\
    \angle BDC &= \arccos\!\left( \frac{l_4^2 + a_{f,2}^2 - l_5^2}{2 l_4 \, a_{f,2}} \right)
\end{align}
\vspace{-1.2em}
\begin{align}
    q_{f,3} &= \angle ADC - \angle BDC
\end{align}}

The resultant position of each finger in its base frame is 
\begin{align}
& x^{f,base}_{f,tip} = {\text c}_0 ((\bp_{f,1}^{f,0})_x + a_1 {\text c}_{123} + a_{f,2} {\text c}_{12} + (a_{f,1} + l_{f,5}) {\text c}_1) \\
& y^{f,base}_{f,tip} = (\bp_{f,1}^{f,0})_y+l_{f,2} {\text s}_{123} + a_{f,3} {\text s}_{12} + (a_{f,1} + l_{f,5}) {\text s}_1  \\
& z^{f,base}_{f,tip} = -{\text s}_0 ((\bp_{f,1}^{f,0})_x + a_1 {\text c}_{123} + a_{f,2} {\text c}_{12} + (a_{f,1} + l_{f,5}) {\text c}_1) 
\end{align}
\noindent with shorthand notation used, e.g., ${\text s}_{0} = \sin(q_{f,0})$ and ${\text c}_{123} = \cos(q_{f,1}+q_{f,2}+q_{f,3})$. 

The thumb ($f=\text{T}$) has an additional passive joint providing affordance for opposition, with the modified transformation:
\begin{equation}
\bT^{T,base}_{T,tip}(\ve{q}^{T},\ve{a}^{T}) =
\bT^{T,base}_{T,op}~\bT^{T,op}_{T,0}~\bT^{T,0}_{T,1}~\bT^{T,1}_{T,2}~\bT^{T,2}_{T,3}~\bT^{T,3}_{T,tip}
\end{equation}
\noindent where $\bT^{T,base}_{T,op}=\begin{bmatrix}
    \mathbf{R}_z(q_{T,op}) & \bp_{op}^{base}) \\ \mathbf{0}_{1\times 3} & 1
\end{bmatrix}$ describes the additional 
\vspace{6pt}
link transform, and then
$\bT^{T,op}_{T,0}=\begin{bmatrix}
    \bR_y(q_{T,0}) & \mathbf{0}_{3\times 1}\\ \mathbf{0}_{1\times 3} & 1
\end{bmatrix}$
followed by the same transforms.
%resulting in the thumb tip position at
% \begin{align}
% x^{base}_{tip} & = (\bp_{op}^{base})_x + (a_1 + l_5)(c_{op}c_1 - s_{op}s_1) + a_2(c_{op}c_{12} - s_{op}s_{12})\notag\\&\quad + c_{op}c_0(\bp_{1}^{0})_x - s_{op}(\bp_{1}^{0})_y + a_3(c_{op}c_{123} - s_{op}s_{123})\\
% y^{base}_{tip} & = (\bp_{op}^{base})_y + (a_1 + l_5)(s_{op}c_1 + c_{op}s_1) + a_2(s_{op}c_{12} + c_{op}s_{12})\notag\\&\quad + s_{op}c_0(\bp_{1}^{0})_x + c_{op}(\bp_{1}^{0})_y + a_3(s_{op}c_{123} + c_{op}s_{123})\\
% z^{base}_{tip} & = -{\text s}_0 ((\bp_{1}^{0})_x + a_1 {\text c}_{123} + a_{2} {\text c}_{12} + (a_{1} + l_{5}) {\text c}_1) + (\bp_{op}^{base})_z
% \end{align}

Given the pose of the glove in the world $\bT^w_{base}$, as measured by an optical tracker, we can derive the pose (position and orientation) of each finger in the world via
\begin{equation}
    \bT^w_{f,tip}=\bT^w_{base}~\bT^{base}_{f,base}~\bT^{f,base}_{f,tip} = 
    \begin{bmatrix}
    \bR_{f,tip}^w & \bp^{w}_{f,tip} \\ \mathbf{0}_{1\times 3} & 1
\end{bmatrix}
\end{equation}
The finger tip positions ($f=\{T, I,M\}$) in the world are thus
\begin{equation}
\bp^{w}_{f,tip} =
\begin{bmatrix}
x^{w}_{f,tip} & y^{w}_{f,tip} & z^{w}_{f,tip}
\end{bmatrix}^\top
\end{equation}

\subsection{Multi-digit Haptic Rendering}

% \begin{figure*}[t]
%     \centering
%     \vspace{2mm}

%     \setlength{\tabcolsep}{6pt} % horizontal spacing
%     \renewcommand{\arraystretch}{1}

%     \begin{tabular}{ccc}
%         \includegraphics[height=1.5in,keepaspectratio]{Figures/force_calibration_setup_2.png} &
%         \includegraphics[width=2.15in,keepaspectratio]{Figures/new_indiv_force.png} &
%         \includegraphics[height=2in,keepaspectratio]{Figures/new_circle_force.png}
%     \end{tabular}

%     \caption{Experimental force tracking results. (Left) Force calibration setup showing the world coordinate frame $(x_{world}, z_{world})$ and tip coordinate frame $(x_{tip}, y_{tip})$. (Center) Individual commanded and measured force components along the $x$ and $z$ axes. (Right) Measured forces plotted in the $x$–$z$ plane compared to the commanded circular trajectory.}
%     \label{fig:force_results}
%     \vspace{-1.5em}
% \end{figure*}

\begin{figure*}[t]
    \centering
    \vspace{2mm}

    \setlength{\tabcolsep}{0pt} % horizontal spacing
    \renewcommand{\arraystretch}{1}

    \begin{tabular}{ccc}
        \includegraphics[height=1.5in,keepaspectratio]{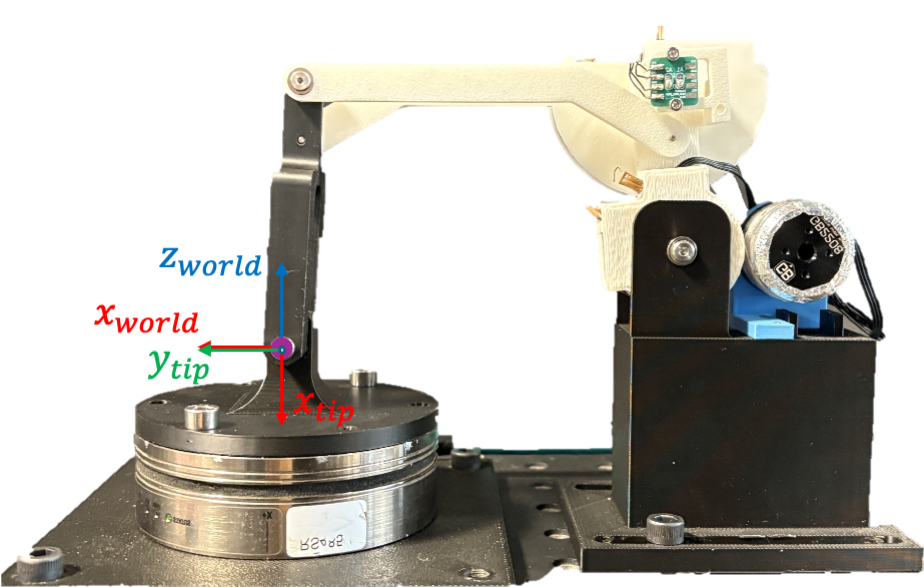} &
        \includegraphics[width=4.5in,keepaspectratio]{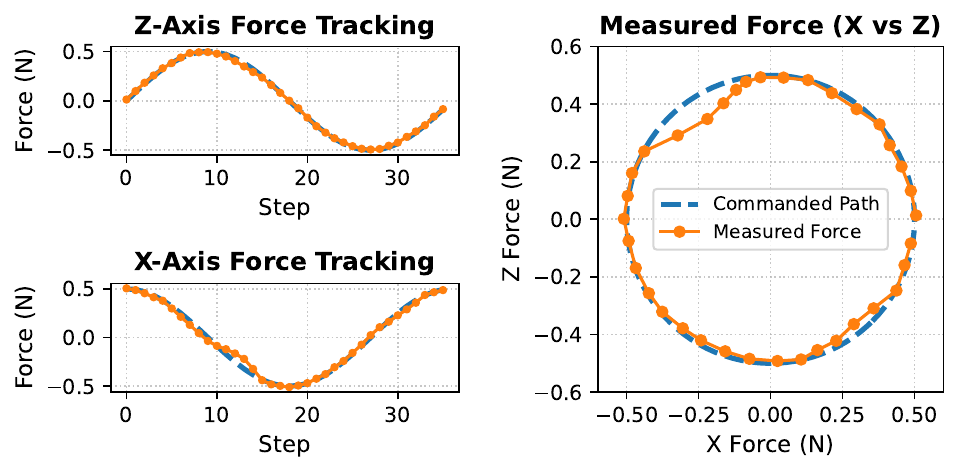} &
    \end{tabular}

    \caption{Experimental force tracking results. (Left) Force calibration setup showing the world coordinate frame $(x_{world}, z_{world})$ and tip coordinate frame $(x_{tip}, y_{tip})$. (Center) Individual commanded and measured force components along the $x$ and $z$ axes. (Right) Measured forces plotted in the $x$–$z$ plane compared to the commanded circular trajectory.}
    \label{fig:force_results}
    \vspace{-1.5em}
\end{figure*}

Directional haptics are rendered by mapping Cartesian forces in the world frame, $ \ve{f}_f^w\in\mathbb{R}^3$, to joint torques $\ve{\tau}_f\in \mathbb{R}^2$ via Jacobian transpose:
\begin{equation}
    \ve\tau_f =\mathbf{J}^w_f(\ve{q}_f)^\top \ve{f}_f^w\label{eq:jacob}
\end{equation}
where $\mathbf{J}^w_f(\ve{q}_f) = \left[\frac{\partial\bp^w_f}{\partial q_{f,1}}, \frac{\partial\bp^w_f}{\partial q_{f,2}}\right] \in \mathbb{R}^{3\times 2}$ is the Jacobian matrix of the position vector $\bp_f^w$ w.r.t the joint angles $\ve{q}_f$. This formulation assumes quasi-static operation, neglecting link inertia and friction, as force transmission is geometry-dominated.

In teleoperation experiments, contact forces are measured directly as 3D Cartesian vectors in the world frame and applied without additional frame transformations. Only translational force components are rendered; moments and wrench relocation are not considered

Because each finger has only two active DOF, the set of achievable Cartesian forces lies in the column space of $\mathbf{J}^w_f(\ve{q}_f)$, whose columns correspond to the instantaneous Cartesian force directions generated by the two motors and thus define a configuration-dependent 2D subspace of $\R^3$. Any desired force can be decomposed as
\begin{equation}
    \ve{f}_f^w = \ve{f}_f^{w,proj} + \ve{f}_f^{w,\perp}
\end{equation}
where $\ve{f}_f^{w,proj}\in\text{col}(\mathbf{J}^w_f)$ and $ \ve{f}_f^{w,\perp}\in\text{null}({\mathbf{J}^w_f}^\top)$.
Since ${\mathbf{J}_f^w}^\top \ve{f}_f^{w,\perp}=0$, this Jacobian transpose mapping eliminates the component out-of the plane, and the device implicitly renders the desired force onto its achievable 2D force subspace. The uncontrolled direction, corresponding to finger adduction and abduction, is therefore not reflected.

\section{Experiments and Results}

We conducted benchtop validation experiments along with a teleoperation user study to provide both quantitative and qualitative assessments of the N2D Haptic Glove, demonstrating its performance in controlled force rendering tasks as well as its effectiveness in real-world robotic teleoperation. 

\subsection{Validation Experiments}

\subsubsection{Planar Joint Actuation Experiment}

A voltage--torque mapping for each joint is identified through a quasi-static directional calibration procedure. Motor torque is proportional to current; at the low velocities considered, back-EMF is negligible, allowing current to be approximated from applied voltage and internal motor resistance via Ohm’s law. 

Planar fingertip forces are measured using an ATI Axia80 3-DOF force sensor. The finger linkage system was mounted at a chosen configuration where the finger tip coordinate axes coincides with the force sensor's measurement axes, as shown in Fig.~\ref{fig:force_results}. All commanded and measured forces are expressed in the world frame defined in the setup.

% AI improved version
% Fingertip forces are measured using an ATI Axia80 force sensor mounted so that the fingertip coordinate axes align with the sensor's measurement axes, as shown in Fig.~\ref{fig:force_results}. All commanded and measured forces are expressed in the world frame defined by the experimental setup. %

Joint torque is modeled as a voltage-dependent mapping
\begin{equation}
{\bf \tau}_f =
\begin{bmatrix}
\zeta_1\, f_{\text{poly}}(v_{f,1},{\bf k}_{f,1}) & 0 \\
0 & \zeta_2\, f_{\text{poly}}(v_{f,2},{\bf k}_{f,2})
\end{bmatrix}
\end{equation}
where $\zeta_i$ are capstan gear ratios and ${\bf k}_{f,i}$ are polynomial coefficients. Using voltage sweeps ${\bf v}_f$ and measured Cartesian forces ${\bf f}_f^w$, the coefficients are solved in closed form via least squares using the Jacobian transpose relation in Eq.~\ref{eq:jacob}. Each finger is calibrated once.

% only use if we fix force calibration w/  linear volt-torq
% Although higher-order polynomials were evaluated to capture potential nonlinearities (e.g., friction, transmission compliance, or motor saturation), a first-order model produced the most stable and geometrically consistent reconstructions. Higher-order fits distorted circular force sweeps (e.g., oblong or skewed trajectories), indicating overfitting rather than true system nonlinearity. A linear model was therefore adopted.

To evaluate tracking accuracy, circular planar force trajectories were commanded. Figure~\ref{fig:force_results} shows the commanded and measured force components $f_x$ and $f_z$ per step.

Force tracking performance is summarized by the root-mean-squared errors (RMSE):
$\mathrm{RMSE}_{x}=0.032~\text{N}$,
$\mathrm{RMSE}_{z}=0.014~\text{N}$,
$\mathrm{RMSE}_{|e|}=0.034~\text{N}$,
and $\mathrm{RMSE}_{\theta}=2.81^\circ$.
The commanded and measured force trajectories show close agreement in both magnitude and direction.

The fingertip Jacobian $\mathbf{J}_f^w$ is well-conditioned with $\kappa(\mathbf{J}_f^w) \approx 2.93$, indicating moderate directional sensitivity in torque-to-force transmission. The Jacobian remains full rank across all evaluated configurations, confirming that independent force components can be generated along both planar axes. These results demonstrate that forces can be rendered accurately throughout the $x$--$z$ plane with minimal distortion.

Residual error is attributed to unmodeled nonlinearities in the voltage-torque mapping, particularly along directions where both motor axes contribute near-equally and commanded voltages are low, entering dead-zone regions. Capstan friction and motor saturation effects further contribute to magnitude and angular discrepancies.

\subsubsection{Directional Force Generation Capability Analysis}
% To ensure N2D Haptic Glove's effectiveness for a wide range of tasks at different hand configurations, its ability to generate forces in different planar directions at various configurations in the workspace (defined as the range of motion of a user's hand while wearing N2D Haptic Glove) must be confirmed. To characterize the directional force capacity of the glove across configurations, we compute the manipulability index

To ensure the N2D Haptic Glove remains effective across different hand configurations, its ability to generate planar forces throughout the workspace must be evaluated. The workspace is defined here as the range of finger configurations achievable while wearing the glove. To characterize planar directional force capability across these configurations, we compute the manipulability index.

    \begin{equation}
        w(\ve{q_f})=\sqrt{\text{det}(\mathbf{J^w_{f,xz}}(\ve{q_f})\mathbf{J^w_{f,xz}}^\top(\ve{q_f}))}=\left|\det\!\left(\mathbf{J^w_{f,xz}}(\ve{q_f})\right)\right|
    \end{equation}
% for all achievable configurations, where $w$ is a scalar metric quantifying the system's ability to generate forces in multiple directions at a given configuration. A higher value indicates that the system can generate forces more effectively and more uniformly in different directions, while $w(\ve{q_f})\rightarrow 0$ indicates proximity to a singular configuration and reduced force controllability. 

for all achievable configurations, where $\mathbf{J^w_{f,xz}}(\ve{q_f}) \in \mathbb{R}^{2\times 2}$ is the planar fingertip Jacobian obtained from the \textit{x-z} rows of the full Jacobian. The value $w$ is a scalar metric that quantifies the system’s ability to generate planar forces in multiple directions at a given configuration. Higher values indicate more uniform and effective directional force generation, whereas $w(\ve{q}_f)\rightarrow 0$ indicates proximity to a singular configuration and reduced force controllability.

% Fig.~\ref{fig:experiments_range_of_motion} shows the manipulability color map for two participants representing small and large hand sizes (15.4 cm and 20.5 cm hand length, respectively, within reported adult ranges \cite{Gordon1989ANSUR}) for their workspace. To obtain all workspaces linkage system joint configurations, the two participants performed maximal flexion, extension, abduction, and adduction movements with the glove on.

%Fig.~\ref{fig:experiments_range_of_motion} shows the resulting manipulability color maps for two participants representing small and large hand sizes (15.4 cm and 20.5 cm hand length, respectively, within reported adult ranges \cite{Gordon1989ANSUR}). To obtain the workspace of achievable linkage configurations, participants performed maximal flexion, extension, abduction, and adduction movements while wearing the glove.%

%By normalizing all manipulability index values, it is shown the index varies moderately ($\approx$30--60\%) in common hand poses for these two participants, indicating that the glove maintains well-conditioned force generation across the functional range of finger motion. As these participants represent the lower and upper end of hand sizes, it is reasonable to extend these results to other hand sizes.%

\begin{figure}
    \centering
    \vspace{2mm}
    \includegraphics[width=1\linewidth]{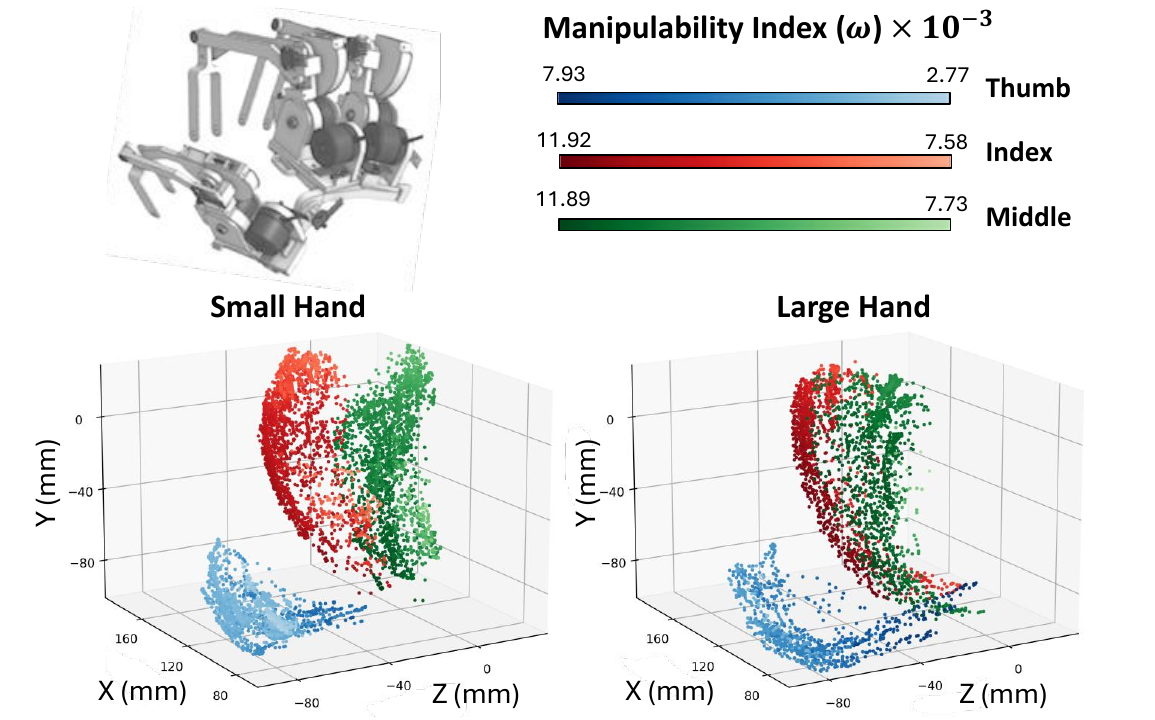}
    \caption{
    Workspace range of motion for a small and large hand (scatter points). The overlaid manipulability colormap indicates configuration-dependent directional force capability, showing well-conditioned force generation throughout typical operating regions.
    }
    \label{fig:experiments_range_of_motion}
    \vspace{-1.5em}
\end{figure}

Fig.~\ref{fig:experiments_range_of_motion} shows the resulting manipulability color maps for two participants representing small and large hand sizes (15.4 cm and 20.5 cm hand length, respectively, within reported adult ranges \cite{Gordon1989ANSUR}). To obtain the configuration space, participants performed maximal flexion, extension, abduction, and adduction movements.

After normalization, 58\% of sampled configurations have manipulability values at or above 0.82. The configuration used in the planar joint actuation experiment also has normalized manipulability 0.82, indicating that over half of the sampled workspace is at least as well conditioned for planar force generation as the experimentally validated configuration. This result suggests that the glove can provide effective directional force output across a broad range of reachable finger postures, rather than only near a small set of favorable configurations for different hand sizes.

\subsubsection{Haptic Teleoperation with Franka Arm Experiment}

% OLD TEXT (Logan)
% Pushing buttons, while seemingly trivial, presents a challenge for teleoperation scenarios without proper haptic feedback. The push task was selected because it captures a second essential manipulation primitive: probing and confirming discrete contacts. In teleoperated environments such as operating control panels or medical equipment, an accurate pushing force is critical. Vision alone cannot always confirm contact, especially when buttons have small travel distances or provide little visual motion, and 1D haptic cues may be insensitive at oblique or out-of-axis misalignments.

%Although seemingly trivial, pushing actions are a fundamental component of manipulation, particularly for probing and confirming discrete contacts, such as pressing buttons. However, in teleoperated environments, including operating control panels or medical equipment, pushing becomes challenging without proper haptic feedback. Vision alone cannot reliably confirm contact, especially when buttons have minimal visual motion, and 1D haptic feedback may fail to capture forces from oblique or out-of-axis misalignment. Therefore, accurate multi-DOF force feedback is critical for naturally and successfully performing precise pushing tasks.

Although seemingly simple, pushing is fundamental to manipulation, particularly for probing and confirming discrete contacts such as button presses. In teleoperated settings—e.g., operating control panels or medical equipment—pushing becomes difficult without appropriate haptic feedback. Vision alone cannot reliably confirm contact when visual motion is minimal, and 1D force feedback fails to capture off-axis interactions. Thus, accurate multi-DOF force feedback is essential for natural execution of pushing tasks.

To evaluate the glove’s ability to render realistic forces during pushing, we conducted a user study simulating teleoperated button pressing. Participants teleoperated a Franka Emika Panda robot equipped with an RH56DFTP dexterous hand with tactile sensing array to press on a digital scale toward a target weight while wearing the N2D Haptic Glove. Three feedback conditions were tested: no haptics, 1D haptics (one motor disabled to emulate a 1-DOF resistive tendon glove), and full 2D haptics.

\begin{table*}
    \vspace{1.5mm}
    \centering
    \caption{\fontsize{9pt}{10pt}\selectfont
    Per-trial absolute error statistics across target weight and contact conditions ($n=16$ participants). Median absolute errors (g) per condition decreases with more degrees of haptic feedback for all groups. $p$-values are from planned pairwise contrasts of a linear mixed-effects model with Holm correction. Negative estimated differences indicate reduced error relative to the comparison modality. From these results, 2D haptic feedback generally yields significantly less error.}
    \label{tab:performance_eval}
    \vspace{0em}
    \renewcommand{\arraystretch}{0.6}
    \setlength{\tabcolsep}{6pt}

    \small
    \begin{tabular}{llcccccc}
        \toprule
        Group & Mode &
        Median Abs Err (Q1–Q3) (g) &
        $p$ (vs Vis) &
        Est. Diff vs Vis (g) &
        $p$ (vs 1D) &
        Est. Diff vs 1D (g) \\
        \midrule

        \multirow{3}{*}{Axial 50g}
        & Visual & 37.68 (20.20-63.00) & -- & -- & $\mathbf{2.53\times10^{-4}}$ & +39.40 \\
        & 1D     & 30.80 (12.28-50.00) & $\mathbf{2.53\times10^{-4}}$ & $-39.40$ & -- & -- \\
        & 2D     & 24.99 (11.25-38.23) & $\mathbf{2.56\times10^{-6}}$ & $-50.60$ & 0.276 & $-11.20$ \\
        \midrule

        \multirow{3}{*}{Axial 100g}
        & Visual & 50.25 (29.88-93.91) & -- & -- & \textbf{0.032} & +26.18 \\
        & 1D     & 34.73 (10.00-87.45) & \textbf{0.032} & $-26.18$ & -- & -- \\
        & 2D     & 21.20 (9.20-43.63) & $\mathbf{1.43\times10^{-5}}$ & $-55.27$ & \textbf{0.032} & $-29.09$ \\
        \midrule

        \multirow{3}{*}{Transverse 50g}
        & Visual & 42.52 (25.00-88.25) & -- & -- & $\mathbf{4.09\times10^{-4}}$ & +36.43 \\
        & 1D     & 27.65 (14.53-50.00) & $\mathbf{4.09\times10^{-4}}$ & $-36.43$ & -- & -- \\
        & 2D     & 27.85 (13.24-47.43) & $\mathbf{5.07\times10^{-4}}$ & $-34.94$ & 0.876 & +1.49 \\
        \midrule

        \multirow{3}{*}{Transverse 100g}
        & Visual & 89.30 (57.00-179.08) & -- & -- & \textbf{0.0315} & +49.02 \\
        & 1D     & 61.00 (27.10-116.59) & \textbf{0.0315} & $-49.02$ & -- & -- \\
        & 2D     & 46.52 (15.88-113.25) & \textbf{0.0315} & $-49.40$ & 0.985 & $-0.37$ \\
        \bottomrule
        
    \end{tabular}
    
    \vspace{-1.5em}
\end{table*}

In all conditions, participants received visual feedback via a live video stream of the robot hand and scale to limit depth cues. The scale display was hidden to replicate typical VR and teleoperation constraints.

%All participants completed the same task for four groups based on robot finger contact location and target weight, with the order of groups being quasi-randomized by the user study organizers. Finger contact location varied between axial probing (pushing with the finger tip) and transverse probing (pushing with the finger pad); target weight varied between 50 grams and 100 grams, forming four group combination. Within each group, participants were tasked to press the scale with the target weight under all three haptic modes, where the order was similarly quasi-randomized by user study organizers.

All participants pressed the scale to a target weight under all three haptic modes across four conditions defined by finger contact orientation and target weight, with group order quasi-randomized. Contact orientation included axial probing (finger tip contact) and transverse probing (finger pad contact), while target weight was set to 50 g or 100 g, yielding four condition combinations.

%Before testing of a group and haptic mode, participants received 2 minutes of training with the corresponding haptic feedback mode active to calibrate their visual and force perception. During this training period, the display screen and scale's reading were shown. After training, participants were asked to press the scale, by lowering their hand, with their perceived target weight for five trials to minimize trial-to-trial variability. For each trial, the true scale reading was recorded by the user study organizer.

Before each condition and haptic mode, participants completed a 2-minute training period with the corresponding feedback active to calibrate visual and force perception. During training, the scale display was visible. Afterward, participants performed five trials in which they pressed the scale—by lowering their hand—to what they perceived as the target weight. The true scale reading was recorded for each trial by the study organizer.

\begin{figure}
    \vspace{1em}
    \centering
    \includegraphics[width=1\linewidth]{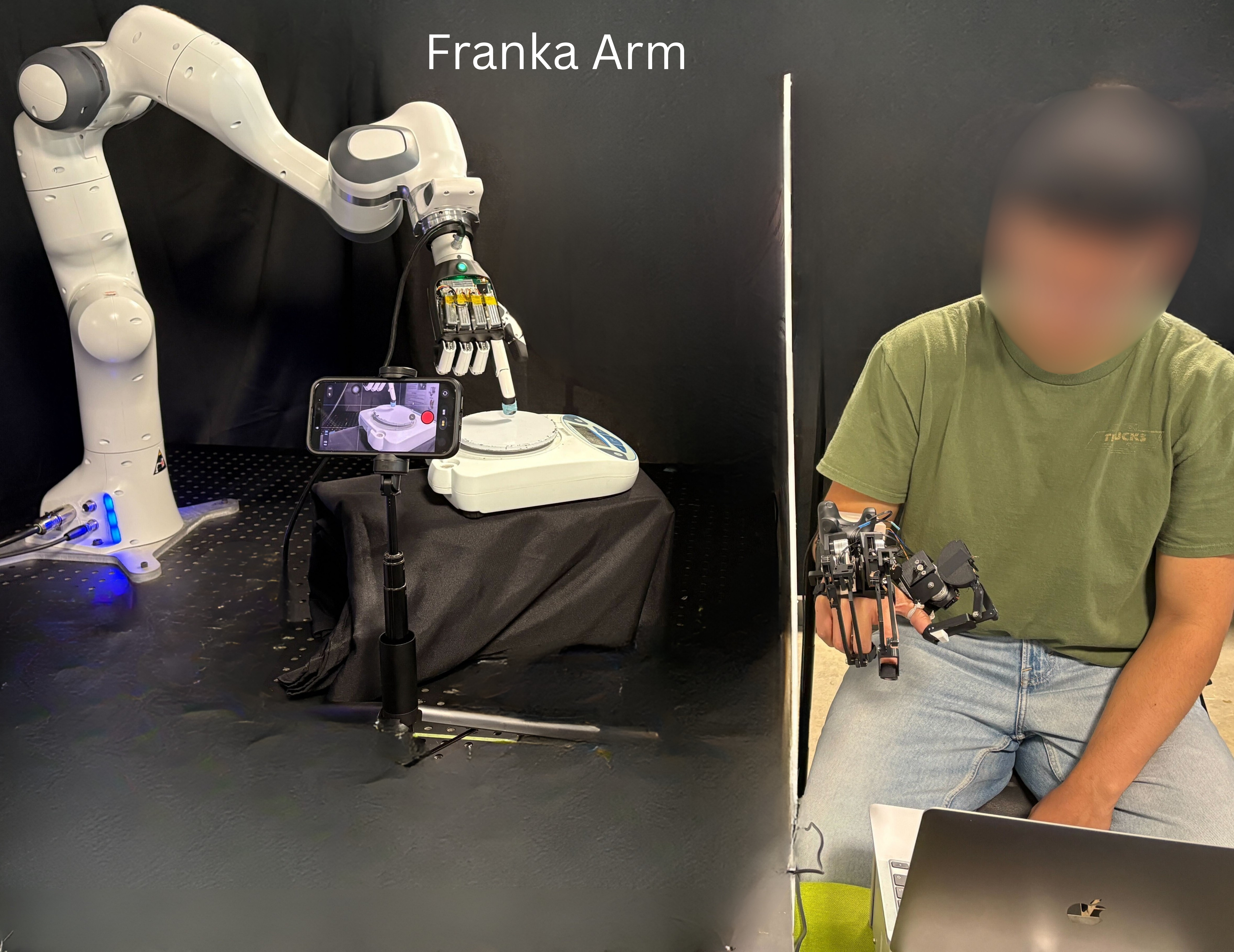}
    \caption{Franka Arm and Inspire Hand scale pressing task setup. Participants teleoperated the robot to press a digital scale toward a target weight under three feedback conditions: no haptics, 1D haptics, and 2D haptics.}
    \label{fig:scale_experiment}
    \vspace{-1.5em}
\end{figure}

% Before testing of a group and haptic mode, participants received 2 minutes of training to calibrate their perception by pressing the scale directly with their finger to experience the target levels for 50 grams and 100 grams of force.

Performance was evaluated by the absolute error between the true scale weight and the target weight. Furthermore, users completed two NASA-TLX~\cite{hart2006nasa} surveys afterwards, one for each finger contact location experience.

% --- Start of Previous Results Section --- 
% \textbf{A NASA TLX survey was conducted for each participant.}
% \begin{itemize}
%     \item Results
% \end{itemize}
% --- End of Previous Results Section --- 

% --- New Results Section --- 
%\subsection{Overall Results}

% OLD TEXT (Logan)
% Figure~\ref{fig:probing_results} shows the results of the user study. In both axial and transverse probing, participants in the no-haptics condition consistently overshot the target ranges, pressing harder than required due to reliance on vision alone. With 1D haptics, participants showed moderate improvement but continued to struggle with precise regulation, particularly at medium and heavy force levels. By contrast, 2D haptics produced the most accurate and stable performance, reducing overshoot and undershoot while lowering trial-to-trial variability. Subjective feedback aligned with these trends as NASA TLX ratings indicated that participants perceived lower workload, reduced frustration, and higher control and confidence with 2D haptics compared to either 1D haptics or no haptics as seen in Figure~\ref{fig:survery_results}. 

\begin{figure}[t]
    \centering
    \vspace{2mm}
    \includegraphics[width=0.95\linewidth]{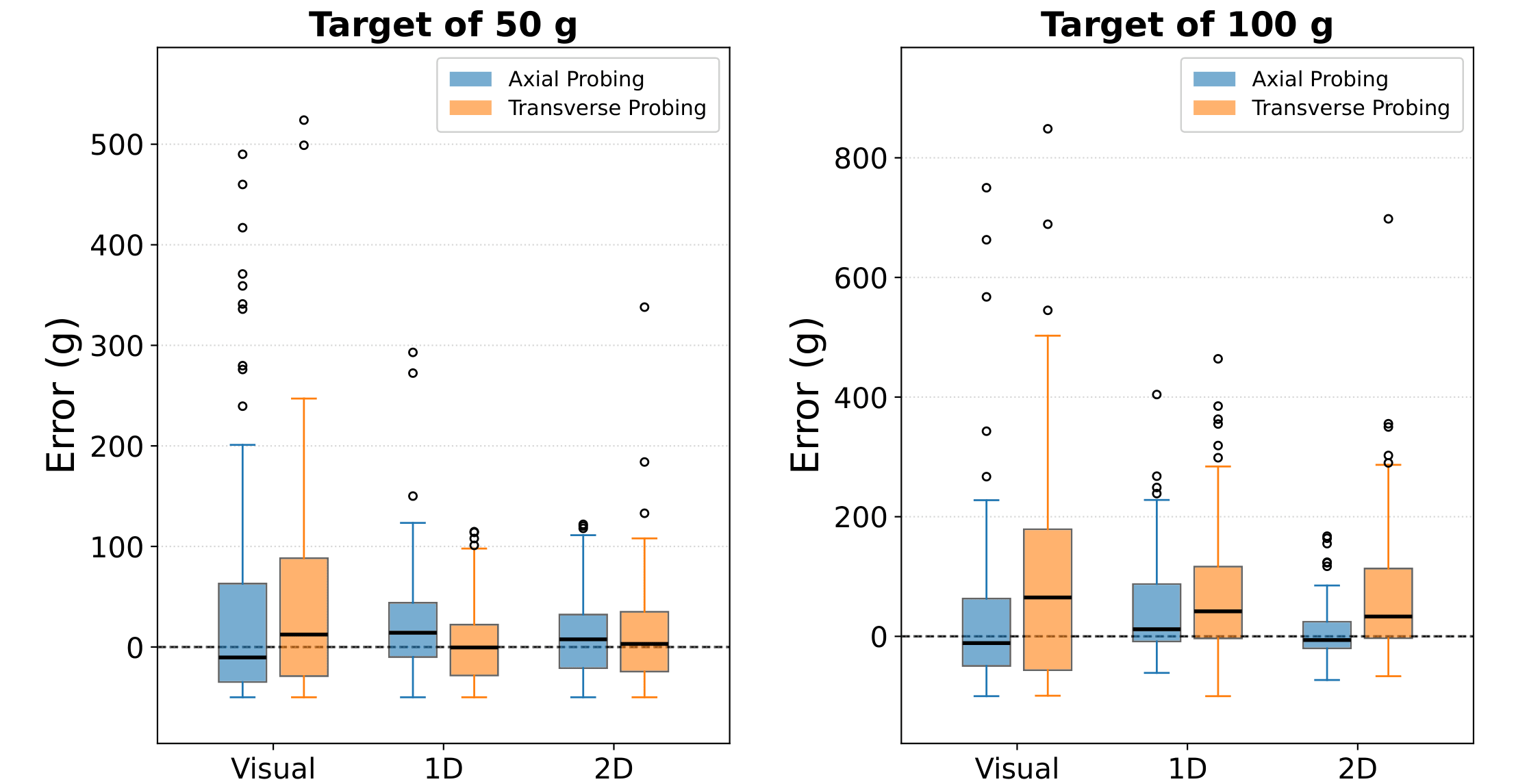}
    \caption{Boxplots of absolute error (g) for target weights of 50 g and 100 g. In axial probing, median error decreases progressively from visual-only to 1D to 2D feedback. In transverse probing, both 1D and 2D feedback reduce error compared to visual-only, with no significant difference between 1D and 2D.}
    \label{fig:probing_results}
    \vspace{-1.5em}
\end{figure}

A total of 16 participants completed the user study, with Figure~\ref{fig:probing_results} showing the absolute error boxplot distribution and Table~\ref{tab:performance_eval} presenting the statistical analysis of all four groups across haptic feedback modes.

To evaluate whether haptic feedback modality, target weight, and finger contact location statistically affect absolute error from target, a linear mixed-effects model (LMM) was used. Absolute error was treated as the dependent variable, with haptic modality, target weight, and contact location as fixed effects. Additionally, participant was included as a random intercept to account for participant baseline differences and the experiment's repeated-measures structure. 

An LMM was chosen, as opposed to Friedman or Kruskal-Wallis with post-hoc tests, due to non-independence of repeated measurements from a participant across trials and groups. Therefore, an LMM allows all individual trial data to be retained while appropriately accounting for within-participant correlations, maintaining statistical power. Planned pairwise contrasts between haptic feedback modalities were performed using Holm-corrected p-values to control for multiple comparisons, with the significance level chosen as $\alpha = 0.05$.

\begin{figure}[t]
    \centering
    \vspace{2mm}
    \includegraphics[width=0.95\linewidth]{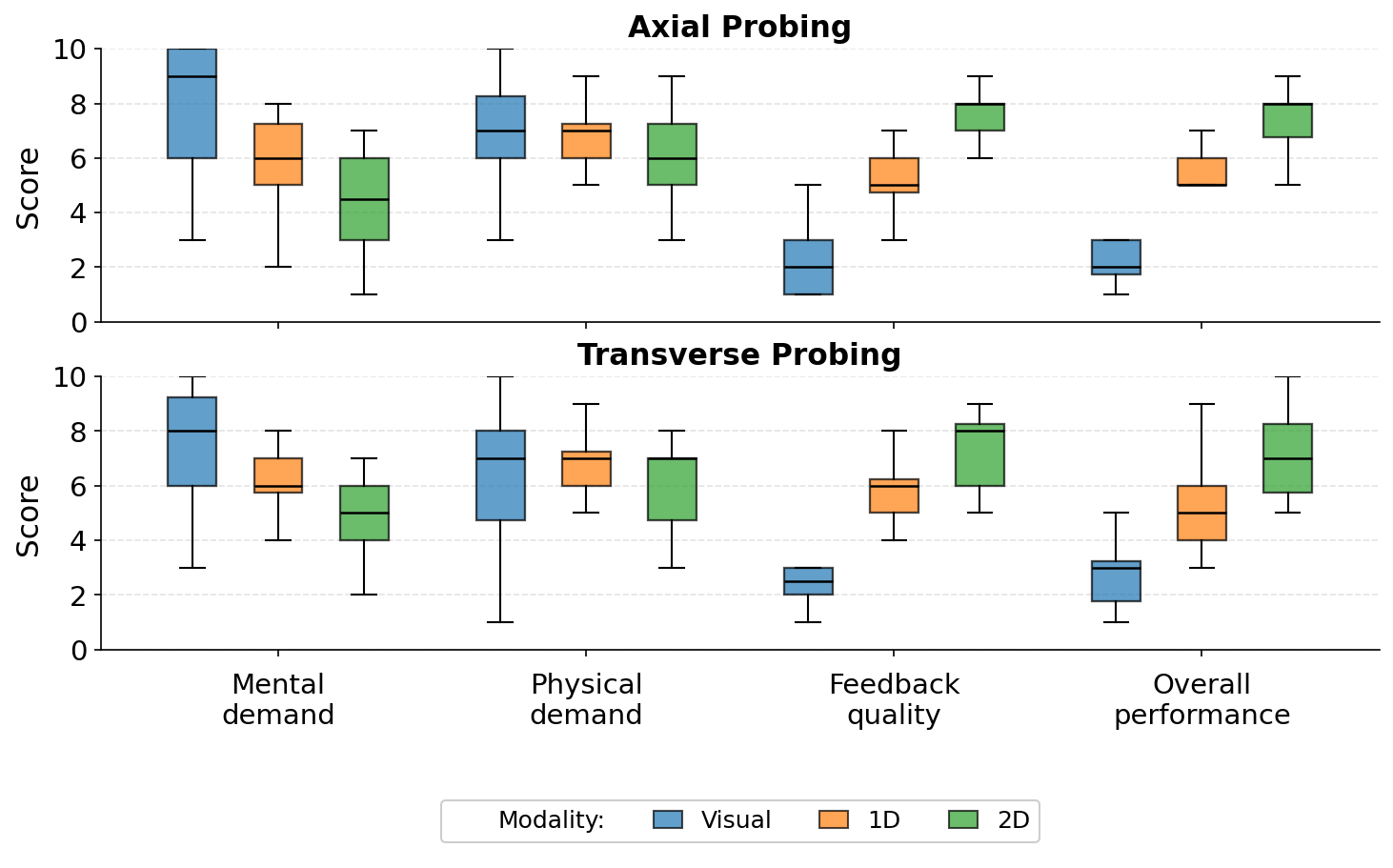}
    \caption{Individual NASA TLX survey results for axial probing and transverse probing, demonstrating multi-directional haptic feedback significantly reduces cognitive load while improving overall performance for participants.}
    \label{fig:survery_results}
    \vspace{-1.5em}
\end{figure}

% In the axial probing groups, median absolute error decreased along no haptic to 1D to 2D haptic, suggesting participants' accuracy improved with more precise haptic feedback. Statistical results from the LMM further support this claim, as at a target of 100 g, receiving 2D haptic feedback was significantly better than receiving both 1D and no haptic feedback, with large estimated differences in favor of 2D feedback. These results indicate 2D haptic feedback is critical in reducing errors and increasing manipulability precision when axially probing due to more accurate force feedback rendering compared to the 1D feedback of resistive tendon gloves. At a target of 50 g, the improvement in median absolute error from 1D to 2D feedback is not statistically significant due to the target weight of 50 g being at the lower end of our motors' working range. Additionally, 2D haptic feedback yields more consistent performance, as reflected by a smaller interquartile range (IQR) of absolute error compared to visual-only and 1D feedback.

In axial probing conditions, median absolute error decreased progressively from no haptic to 1D to 2D feedback, indicating improved accuracy with higher-fidelity force rendering. The LMM results corroborate this trend: at the 100 g target, 2D feedback significantly outperformed both 1D and no-feedback conditions, with large estimated differences favoring 2D. These findings demonstrate that 2D haptic feedback significantly enhances force regulation and precision during axial interactions, where single-DOF resistive tendon feedback is insufficient.

At the 50 g target, the reduction in median error from 1D to 2D was not statistically significant, likely because 50 g approaches the lower bound of the motors’ effective operating range. Notably, 2D feedback produced more consistent performance, reflected by a smaller interquartile range of absolute error compared to visual-only and 1D conditions.

% However, in transverse probing groups, participants when receiving 2D and 1D haptic feedback performed significantly better compared to receiving no haptic feedback. However, no statistically significant difference was detected between receiving 2D and 1D haptic feedback, with a corresponding near zero estimated difference. These results indicate any directional haptic feedback is essential in improving accuracy and reducing errors when transversely probing, but 2D haptic feedback is not conclusively better or worse compared to 1D feedback. These conclusions align with expected outcomes since force feedback of transverse probing can be accurately rendered from existing 1-DOF resistive tendon force gloves.

In transverse probing conditions, participants performed significantly better with either 1D or 2D haptic feedback compared to no feedback. However, no statistically significant difference was observed between the 1D and 2D modes, with an estimated effect size near zero. These results suggest that the presence of directional force feedback—regardless of dimensionality—is sufficient to improve accuracy and reduce error during transverse interactions, while 2D feedback offers no measurable advantage over 1D. This outcome is expected, as transverse probing forces can be effectively rendered using conventional 1-DOF resistive tendon-based gloves.

Subjective feedback further reinforces the value of the N2D Haptic glove over existing 1D resistive tendon gloves. NASA TLX ratings indicate that when receiving 2D haptic feedback for both axial and transverse probing, participants expressed that their perceived performance increased with more control and clearer feedback, while their mental demand decreased, as seen in Figure~\ref{fig:survery_results}. Moreover, responses indicate total perceived physical demand only decreased marginally with 2D haptic feedback due to the excessive weight of the glove present for all three feedback modes, as expressed by multiple participants.

Taken together, results from our study show that planar force feedback enhances both objective force regulation and subjective user experience in teleoperated tasks.

% --- Start of Previous Discussion and Conclusion Section --- 
% \section{Discussions and Conclusions}
% \textbf{Directional haptic feedback on fingers has significance and importance in achieving realistic physical interaction with virtual envrionments. and teleoperated systems.}
% \begin{itemize}
%     \item Go over the various things achieved by the glove
% \end{itemize}

% \textbf{Designing an haptic glove is a delicate tradeoff of performance of haptic fidelity, and physical comfort.}
% \begin{itemize}
%     \item bring up both benefits and limitations of the design
% \end{itemize}

% \textbf{Haptic N2D presents the first directional haptic feedback in a multi-fingered haptic interface, providing a new capability in realistic force rendering that is important for the VR, robot teleoperation, and imitation learning community.} 
% \begin{itemize}
% \item mention some immediate applications
% \item end off with what, if improved or advanced, could the next glove provide and where it could lead us to.
% \end{itemize}
% --- End of Previous Discussion and Conclusion Section --- 

% --- New Discussion and Conclusion Section --- 
\section{Discussions and Conclusions}

Directional fingertip haptics are essential for realistic physical interaction in teleoperation. We presented the design, analysis, validation, and user study of the N2D Haptic Glove to evaluate the impact of multi-directional feedback. 

Results show that 2D haptic feedback significantly improves precision in axial probing and pressing tasks by reducing force error and increasing consistency. This highlights the feasibility of implementing the N2D haptic glove for VR and teleoperation axial probing tasks, such as pushing small buttons or searching a confined medical environment for surgical tools with finger tips.

Although effectiveness of the N2D Haptic Glove for transverse probing is comparable to 1D haptic feedback, the collective results suggest the glove will still enable more precise manipulation than existing 1D haptic gloves for tasks in between full axial and full transverse probing directions. Additionally, N2D Haptic Glove's 2-DOF force feedback improves user experience and confidence, further proving its value for teleoperation and VR simulation tasks. 

Collectively, these findings demonstrate that multi-directional force feedback allows users to regulate contact forces more accurately and consistently across a broader range of tasks. The N2D Haptic Glove thus provides a foundation for next-generation wearable haptic interfaces that support more natural and effective human–robot interaction.

%However, while these user study results, alongside other benchtop validation tasks, demonstrate our design's ability to accurately render multi-directional forces while preserving user natural range of motion, limitations still exist. 

Although the work demonstrates accurate multi-directional force rendering, several limitations remain. The current design actuates three fingers and weighs 562 g; scaling to all five would increase total mass to approximately 780 g. At the current weight, the glove can become fatiguing during extended use, despite the GB2208 being one of the best torque-to-weight motors on the market. Another limiting factor is the deadzone at the lower bound of the motor's operating voltage, which limits feedback for very precise, low-force tasks. Ongoing advances in drone motor technology are rapidly improving motor size and efficiency, which is likely to reduce these constraints and offer a more effective N3D option in the future.

In conclusion, the N2D Haptic Glove demonstrates that multi-directional haptics are both feasible and beneficial for contact-rich tasks, particularly those involving axial probing. Our user study highlights its advantage in fingertip button pressing, but similar benefits extend to tasks categorized as peg-in-hole insertion, where sensing lateral misalignment forces can substantially improve guidance and task completion time. More broadly, the N2D Haptic Glove may be integrated with immersive simulation environments \cite{gani2022impact,li2024haptic}, used in medical teleoperation systems \cite{guo2024lightweight}, and incorporated into imitation learning pipelines \cite{li2023immersive}. These applications demonstrate the value of multi-directional fingertip feedback as a missing modality in current wearable haptic systems.

\balance
\bibliographystyle{ieeetr}
\bibliography{refs}
\balance

\end{document}